# Balancing Natural Language Processing Accuracy and Normalisation in Extracting Medical Insights

NLP in Extracting Medical Insights

Paulina Tworek*

Personal Health Data Science Team, Sano - Centre for Computational Personalized Medicine, Krakow, Poland

p.tworek@sanoscience.com

Miłosz Bargieł

Personal Health Data Science Team, Sano - Centre for Computational Personalized Medicine, Krakow, Poland,

Faculty of Physics, Astronomy and Applied Computer Science, Jagiellonian University, Krakow, Poland

m.bargiel@sanoscience.com

Yousef Khan

Personal Health Data Science Team, Sano - Centre for Computational Personalized Medicine, Krakow, Poland

y.khan@sanoscience.com

Tomasz Pełech-Pilichowski

Institute of Computer Science, AGH University of Krakow, Krakow, Poland

tomek@agh.edu.pl

Marek Mikołajczyk

Voivodeship Rehabilitation Hospital for Children in Ameryka, Ameryka, Poland

m.mikolajczyk@ameryka.com.pl

Roman Lewandowski

Institute of Management and Quality Sciences, Faculty of Economics, University of Warmia and Mazury, Olsztyn, Poland

roman.lewandowski@uwm.edu.pl

Jose Sousa

Personal Health Data Science Team, Sano - Centre for Computational Personalized Medicine, Krakow, Poland

j.sousa@sanoscience.com

Extracting structured medical insights from unstructured clinical text using Natural Language Processing (NLP) remains an open challenge in healthcare, particularly in non-English contexts where resources are scarce. This study presents a comparative analysis of NLP low-compute rule-based methods and Large Language Models (LLMs) for information extraction from electronic health records (EHR) obtained from the Voivodeship Rehabilitation Hospital for Children in Ameryka, Poland. We evaluate both approaches by extracting patient demographics, clinical findings, and prescribed medications while examining the effects of lack of text normalisation and translation-induced information loss. Results demonstrate that rule-based methods provide higher accuracy in information retrieval tasks, particularly for age and sex extraction. However, LLMs offer greater adaptability and scalability, excelling in drug name recognition. The effectiveness of the LLMs was compared with texts originally in Polish and those translated into English, assessing the impact of translation. These findings highlight the trade-offs between accuracy, normalisation, and computational cost when deploying NLP in healthcare settings. We argue for hybrid approaches that combine the precision of rule-based systems with the adaptability of LLMs, offering a practical path toward more reliable and resource-efficient clinical NLP in real-world hospitals.

## 1 INTRODUCTION

Since the public demonstration of IBM Watson in 2010, Natural Language Processing (NLP) has been heralded as a transformative technology for healthcare [1]. By extracting knowledge from unstructured text, NLP has the potential to enhance clinical decision support, automate administrative tasks, and improve patient outcomes. Yet, despite over a decade of progress, reliable deployment of NLP in clinical practice remains an open challenge [2, 3].

Electronic health records (EHRs) contain large volumes of unstructured text with high variability in style, terminology, and completeness. Inconsistencies such as misspellings, local dialects, and ambiguous abbreviations complicate information retrieval. Standardised terminologies, such as SNOMED-CT and ICD-10, have improved interoperability, but substantial heterogeneity persists. The difficulty is particularly acute in non-English healthcare systems, where models are underdeveloped and domain-specific resources are scarce [4].

Recent advances in Large Language Models (LLMs) have rekindled optimism [5, 6]. These models can summarise, interpret, and extract medical information across multiple languages, going beyond translation-based approaches [7]. However, they remain computationally expensive and prone to hallucinations or inaccuracies, especially in underrepresented languages. In contrast, rule-based methods, although less flexible, are lightweight and sometimes more precise in tasks that require strict adherence to medical syntax [8]. The resulting trade-off between adaptability, accuracy, and computational cost is especially relevant for hospitals with limited IT infrastructure.

In this work, we present a comparative analysis of rule-based and LLM-based methods for extracting structured clinical information from unstructured hospital records in Polish. Using records from pediatric patients with allergic diseases at the Voivodeship Rehabilitation Hospital for Children in Ameryka in Poland, we evaluate the extraction of demographics, clinical findings, and prescribed medications - factors critical for allergy classification and medical decision-making [9-11]. Our research evaluates two distinct approaches for this task that involve different computing power requirements (illustrated in pink within Scheme 1):

1. **A Rule-based approach**, utilizing pattern-based matchers and simple rule-driven methods to extract demographic and clinical data with minimal computational cost.
2. **A LLM-based approach**, utilizing generative AI for more refined text interpretation, though with substantially increased computational requirements.

We also assess how translation from the native language into English influences accuracy, highlighting both benefits and risks, such as the loss of gender cues.

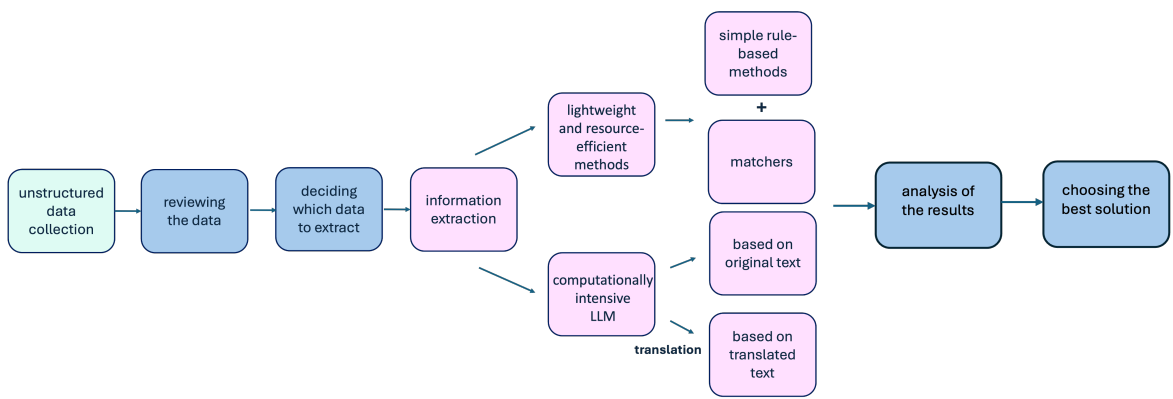

Scheme 1: The workflow for implementing systems for information extraction from medical texts. The steps highlighted in green and blue indicate general implementation, while those in pink represent approach proposed by us.

Our findings show that rule-based systems are highly accurate for demographic extraction, while LLMs outperform at recognising drug names without the need for handcrafted dictionaries. Translation improves specific tasks but reduces effectiveness of others, underscoring the importance of linguistic nuance. By systematically analysing these trade-offs, this work contributes to the ongoing discussion on how best to deploy NLP in real-world healthcare. We argue that hybrid approaches, which combine the precision of rule-based systems with the adaptability of LLMs, offer a practical path toward robust and resource-conscious clinical NLP.

## 2 RELATED WORKS

The effective extraction and analysis of medical data are crucial for enhancing clinical decision support, automating administrative tasks, and improving patient care. Despite advancements in NLP, significant challenges remain in healthcare applications, including information loss, context preservation, normalisation, and domain-specific accuracy. Early approaches relied on rule-based systems supported by ontologies, whereas recent advances in Large Language Models (LLMs) have introduced more adaptive, multilingual methods. However, these two paradigms differ substantially in computational requirements, scalability, and robustness when applied to real-world clinical texts, particularly in non-English contexts. This section reviews prior work on the applications of NLP in medicine, the role and limitations of rule-based methods, and the emerging use of LLMs in healthcare, before positioning our contribution within this landscape.

## 2.1 Applications of NLP in medical context

NLP has been widely applied across healthcare, from clinical note summarisation and automated medical coding to decision support systems and patient-facing applications. Prior studies demonstrate its role in enhancing knowledge discovery from electronic health records (EHRs), supporting clinical trial matching, and enabling conversational agents for healthcare [12-19]. These applications underline the potential of NLP to streamline workflows and improve patient care. However, performance often depends on the availability of standardised text and domain-specific training data, both of which are unevenly distributed across languages and healthcare systems.

## 2.2 Challenges in Rule-Based Medical NLP Systems

Before data-driven machine learning (ML) models took root, rule-based systems were the primary approach in medical information extraction. These systems rely on manually curated rules and structured ontologies, such as SNOMED-CT and ICD-10, to standardize terminology and enhance clinical documentation [20, 21]. Rule-based NLP has been widely used for tasks requiring strict adherence to syntax, such as automated medical coding and domain knowledge. Despite this precision, rule-based systems exhibit several limitations. Their rigid structure makes the less adaptable to linguistic variability, leading to reduced efficacy in handing diverse medical texts. Additionally, these systems require extensive manual development and maintenance, limiting their scalability in multilingual and evolving healthcare environments [8]. While rule-based methods remain relevant in resource-constrained settings, their limitations have spurred the adoption of ML-based NLP techniques.

## 2.3 Large Language Models in Healthcare

Recent advances in Large Language Models (LLMs), including the GPT family, Llama, and Gemma, have demonstrated strong performance in text summarisation, entity recognition, and the multilingual processing of medical records [22]. Their adaptability makes them suitable for heterogeneous EHRs, and studies have shown they can approach or even surpass human experts in some summarisation tasks [7, 23, 24]. However, LLMs also present critical limitations: they may generate hallucinated content, struggle with domain adaptation, and exhibit inconsistent performance in underrepresented languages [25, 26]. Moreover, their computational demands are substantial, posing significant barriers to deployment in hospitals with limited IT resources [27-34].

## 2.4 Comparative Performance and Implementation Challenges

Comparative studies between rule-based systems and LLMs reveal that while LLMs offer better adaptability and scalability, rule-based methods retain an edge in precision-sensitive applications [24, 35, 36]. Yet, there is limited research directly comparing rule-based and LLM-based methods in clinical contexts, especially in non-English real-world healthcare systems [37, 38]. Furthermore, the effects of translation on medical information extraction have received little systematic evaluation, despite their relevance for multilingual deployments [39-40].

## 2.5 Positioning of This Work

Our work addresses these gaps by directly comparing rule-based and LLM-based approaches on unstructured hospital records in a non-English language setting. We evaluate their relative strengths for extracting demographics, clinical findings, and medications, and analyse the impact of translation into English. By doing so, we provide new insights into the trade-offs between accuracy and normalisation, taking into account computational efficiency, and argue for hybrid approaches tailored to real-world healthcare environments.

## 3 MATERIALS AND METHODS

### 3.1 Dataset

Data was collected from epicrisis (medical summaries) comprised of 1,679 pediatric patients diagnosed with allergic diseases at the Voivodeship Rehabilitation Hospital for Children in Ameryka, Poland. Each patient was diagnosed by one of three selected doctors. The data is in form of unstructured text written in Polish. According to the regulations of The Minister of Health in Poland dated November 9, 2015, epicrisis should provide all the necessary information about condition of the patient, symptoms, medical procedures, treatment and recommendations [41]. Before the data was released to us, it was anonymised, and information was additionally checked to prevent the identification of the patients. This sensitive information, like identification number, date and place of birth, parents' names and surnames, is stored in another part of the hospital system and goes to another database. Each patient was diagnosed and assigned a code according to the International Classification of Diseases ICD-10 [42]. As different physicians recorded the data, there may be variations in its presentation. This assumption is also confirmed by doctors in this hospital who do not use templates for epicrisis.

### 3.2 Experiments description

#### 3.2.1. Extraction of information on age and sex of patients

Age and sex are important information from the point of view of child diagnosis. According to the literature, the development of allergies and allergic symptoms is closely related to age, with some primarily affecting infants and often resolving early, while others persist into adulthood. The severity and presentation of allergic reactions also change over time, with infants typically experiencing milder symptoms such as hives and vomiting, while older children and adults are more prone to severe reactions like anaphylaxis and airway involvement [18, 41]. The patient's sex can also influence the development and diagnosis of allergies as it can have specific associations with certain allergens (e.g., males are more likely to develop dust mite and fungal allergies) [42]. After reviewing the data received from the hospital, it was noted that age and sex are only included in epicrises. Therefore, it is necessary to extract this information from the unstructured data. The extracted data must then be transformed into a tabular format. This transformation enables integration with structured medical examination data. The goal is to implement this information in a classification system that supports medical decision-making in the future. A comparison was performed between information retrieval using Matcher rule-based methods and LLMs.

#### 3.2.2. Extraction of information of mentioned medications and skin lesions

Observing skin lesions is crucial for classifying the type of allergy, as information on their presence can significantly narrow down the number of conditions, leading to a more accurate diagnosis and targeted treatment. Additionally, skin lesions may be an early indicator of systemic allergic reactions, guiding further diagnostic tests such as skin prick testing or blood tests for allergen-specific IgE levels [43]. Conversely, details about the medications taken can aid in diagnosis, as certain drugs may either provoke allergic reactions or diminish the disease symptoms, potentially affecting the assessment [44, 45]. This means that both pieces of information are relevant for diagnosing allergies. A performance comparison was conducted between simple rule-based methods and selected LLMs.

### 3.2.3. Comparison of the accuracy of information extraction using LLMs after translation of epicrises from Polish to English

Llama 3 was predominantly trained on English-language data, supplemented by some data from other languages. It is regarded as a multilingual model. Although Meta acknowledges that its performance in non-English languages may not be as strong as in English [46]. Similarly, Gemma 7b was trained on primarily English data from web documents, mathematics, and code, but Google noted that the Gemma family of models possesses some multilingual capabilities; nonetheless, their performance in languages other than English often falls short of optimal [47, 48]. Therefore, we decided to check if the models would perform better on texts translated from Polish into English. Firstly, the epicrises were translated to English. Then, the information extraction process was repeated for the selected LLMs on the translated texts, and the extraction accuracy was then compared for the original texts in Polish and English.

## 3.3 Methods

### 3.3.1. Matcher approach for age and sex extraction

SpaCy Matcher is a Python rule-matching engine featuring regular expression matching [49]. This approach enabled us to find the patient's sex and age by identifying token sequences aligning with pattern rules. SpaCy Matcher enhanced the rule-based approach by incorporating morphological analysis to understand the structure and grammatical features of words in text. Performing this kind of analysis is particularly relevant for analyzing Polish texts, due to their complex grammar, where words change according to attributes like the case, sex, number, and tense [49]. A preliminary review of the text indicated that sex and age are in the first sentence of the epicrisis, specifically within the first six words. Therefore, the analysis in terms of age and sex has been limited to this scope of the text.

### 3.3.2. Simple rule-based approach for drugs and skin lesions extraction

Since many terms describe skin changes, the methods for defining them were first identified based on 70 epicrises per doctor. It was also observed that each patient is assigned only one term. A database of these terms (keywords) consisting of a dozen items was then created (the words included in the database are presented in Supplementary Materials). The elements in the database occurred in a shortened form that was resistant to the grammatical variation of words. Then, each word occurring in the epicrisis was compared with each word in the database representing skin lesions. If a word from the database was matched as a fragment or whole word in an epicrisis, it was considered an identified skin lesion. Accuracy was used as an evaluation metric here.

Drugs: Each word in the epicrisis was compared to each word in the database of drugs sold in Poland. For each pair of words (word in epicrisis-word from the drug database), we determined the sequence similarity based on a modified version of Levenshtein distance [35]. The value was obtained by implementing *token_set_ratio* from the fuzzywuzzy library [50]. Modifying Levenshtein distance involves disregarding duplicates and insensitivity to the order of words. The sequence similarity was used as a threshold for filtration. Pairs with a threshold value equal to or greater than 80 were recognized as entities with the same meaning. A threshold value lower than 100 was adopted, as word variation or typos can cause differences between words with identical meanings. This means that if the threshold was reached, the given word from the epicrisis was identified as the name of the drug.

Drug extraction involved identifying drug names and ensuring that all medications taken by a patient were accurately captured while preventing the inclusion of any drugs not mentioned during the epicrisis. Therefore, the accuracy used in other experiments is not the most suitable. This led to a new metric that considers cases where incorrect drug names have been extracted but also considers incorrect quantities of drugs' names. The metric has been called Adjusted Accuracy, and it is represented by Equation 1:

$$\text{Adjusted Accuracy} = \frac{\sum_{i=1}^{N} \frac{\min(length(y_i),\ length(\hat{y}_i))}{\max(length(y_i),\ length(\hat{y}_i))} + \frac{length(correctfound_i)}{length(tofind_i)}}{2N} \quad (1)$$

Where $y_i$ is the vector of the names of the drugs mentioned in epicrisis, $\hat{y}_i$ is the vector of the extracted name of drug, and $N$ is the number of names of the drugs mentioned in epicrisis for specific patient.

### 3.3.3. Large Language Models (LLM) for age, sex, skin lesions and drugs extraction

The second approach was to apply Large Language Models (LLM) to extract information from epicrises on age, sex, skin lesions and medications taken by patients. The performance of two popular and well-validated models was checked and compared. The Llama-3-8B-it and Gemma-7b-it were selected, as both models are easily available to a wide range of users including researchers and companies and can be used without sending data to external servers [51, 52]:

- Llama-3-8B-it belongs to Meta Llama 3 family of LLMs [51]. Llama 3 is an auto-regressive model using optimized transformer architecture. It was pre-trained on over 15 trillion data tokens from publicly available sources. The 8B-Instruct parameter size version is an instruction-tuned model developed for different types of natural language generation tasks like dialogue use cases. The Llama 3 model is accessible to a broad audience, including researchers and businesses, but its license does not fully align with the traditional definition of open source [46].
- Gemma-7b-it belongs to the Gemma family of lightweight LLMs developed by Google DeepMind and other teams across Google [52]. The model was trained on a dataset containing 6 trillion tokens of text data. The Gemma models are designed for text generation tasks, reasoning, question answering and summarization. It is an open-source model [53].

The structure of the applied prompt was as follows:

$$< key1 = value1 \mid key2 = value2 \mid key3 = value3 \mid ... \mid keyN = valueN >$$

and the following instructions were included in the prompts:
- It is a healthcare assistant designed to extract medical-related information,
- The answer should be given only based on the context given,
- The example of expected output and its format was provided,
- The set of possible keys that should be extracted was presented: age, sex, drugs mentioned and skin changes. All keys should be returned, if any information according to the model is unavailable, then the key should be returned as "*key=None*".

### 3.3.4. Translation

Llama CPP was used to translate the texts from Polish into English [54]. The following instruction has been given to the model:

- Your goal is to translate every sentence from Polish to English.

A manual translation quality assessment was conducted on 70 epicrises for each of the 3 doctors, confirming that the translations were very high quality.

## 4 RESULTS AND DISCUSSION

### 4.1 Calculating basic metrics for doctors' text length to assess unstructured data normalisation

The absence of standardized practices for collecting medical data poses a major obstacle to implementing data-driven healthcare systems. To determine if variations in text format among doctors in our dataset exist, we analysed the average number of sentences and words in the epicrises (Table 1). As it was observed that information about age and sex appeared in the first sentence of each of the three doctors, we also calculated the average word count of this initial sentence for each specialist (Table 1).

Table 1: Comparison of text length statistics for individual doctors: mean and standard deviation for number of sentences, number of words for whole epicrisis and for its first sentence.

| Doctor | Number of sentences | | Number of words | | Number of words in first sentence | |
|---|---|---|---|---|---|---|
| | Mean | Std | Mean | Std | Mean | Std |
| 1 | 8.55 | 2.84 | 81.40 | 23.42 | 18.19 | 7.14 |
| 2 | 14.91 | 3.41 | 146.07 | 36.57 | 14.91 | 3.41 |
| 3 | 17.73 | 4.76 | 258.56 | 63.09 | 30.81 | 14.17 |

The table shows that the selected text metrics vary significantly for each physician. The mean numbers of sentences and words of all the epicrisis are the lowest for Doctor 1 and the highest for Doctor 3. This may be related to sentences containing different amounts of information, text structures, or personal writing styles. These factors indicate a lack of data normalisation, a typical phenomenon in medical units, especially for unstructured data. Additionally, the average standard deviation for the number of sentences and words for all doctors represents approximately 25% of the average value. The standard deviation for the number of words in the first sentence varies from about 23% of an average number of words in the first sentence for Doctor 2 to 45% for Doctor 3, which exhibits the highest variation. The high standard deviation values indicate significant variability in text length among doctors, potentially influenced by factors such as the type of symptoms present and, consequently, the corresponding diagnosis. To test this assumption, the same metrics were also calculated for the four most common diagnoses (J45.0 - predominantly allergic asthma, J45.9 - asthma, unspecified, L20.8 - other atopic dermatitis, and L27.2 - dermatitis due to ingested food and the results are shown in Figure 1 [42].

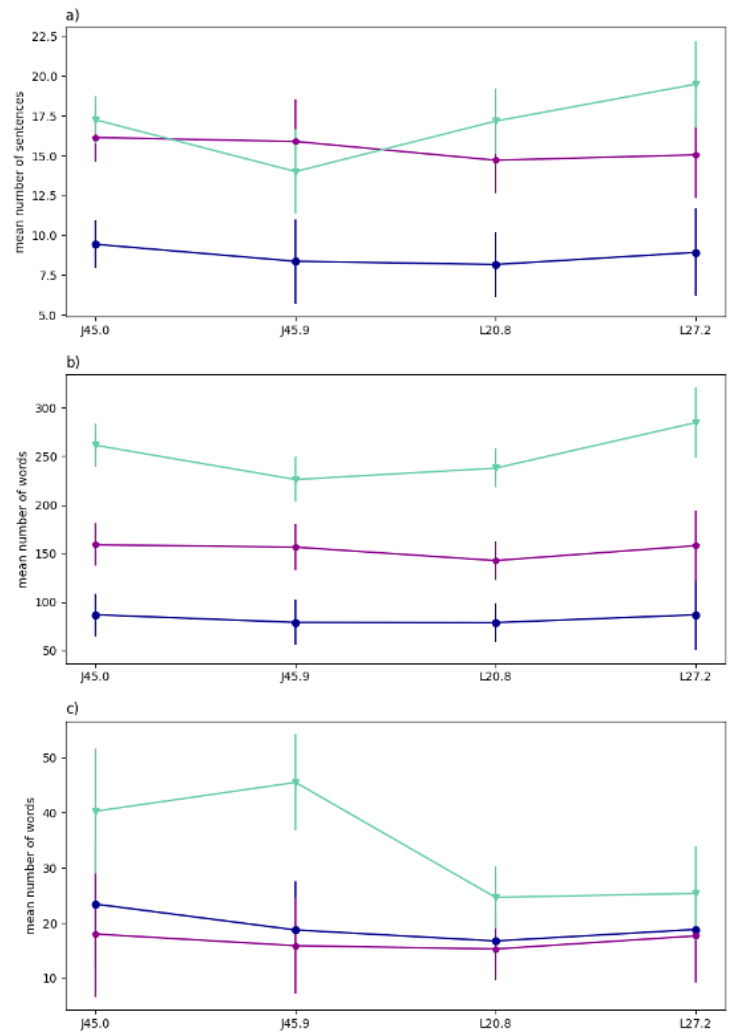

Figure 2: Comparison of text length statistics for individual doctors dependent on the final diagnosis: a) average number of sentences, b) average number of words, c) average number of words in first sentence (Doctor 1 - ●, Doctor 2 - ♠, Doctor 3 - ▲).

The graphs (Figure 1a and Figure 1b) clearly show significant differences in the value for the number of sentences and words in epicrises depending on the diagnosis of Doctor 3. Trends are less distinct for Doctors 1 and 2, but the higher mean values for sentence and word counts in the first sentence of epicrises are observed for J45.0 and J45.9 – both representing different types of asthma. In contrast, the other two codes, L20.8 and L27.2, are linked to dermatitis. A comparison of Doctor 3's trends with those of Doctors 1 and 2 reveals notable differences, further reinforcing the thesis that each doctor has a unique filing style.

## 4.2 Extracting Age and Sex Using Rule-Based Matchers

As stated in subsection 3.2.1, age and sex information is provided in the first sentence regardless of the doctor, simplifying the analysis. After reviewing the epicrises, it has been noted that age is recorded varies between entries. Still, in most cases, it is a natural number or an age format that is completely atypical in everyday language in the form of a natural number followed by an optional word connector and a fraction of 'natural number/12'. It was noted that this format represents the number of the 12 months of the year (e.g., "5 4/12" or "5 i 4/12" means 5 years and 4 months). There were also some reports where the age was represented just by a floating-point number (e.g. 5.5). The conclusions reached during the review of the interviews allowed the creation of basic rules to describe how age was recorded.

It is different in terms of sex, which in most cases was not explicitly mentioned. In this instance, linguistic characteristics can be utilized. Polish is a morphologically rich language, which is reflected in sex-based inflection. The Polish language features grammatical sex (masculine, feminine, and neutral) and their distinction in the plural form (masculine-personal and non-masculine-personal), which affects the inflection of nouns, adjectives, pronouns, and verbs. As a result, word forms often adjust to the sex of the person they refer to, making Polish exceptionally rich in terms of inflection. Therefore, the rule-based matcher method from the SpaCy library has been used as a less computationally intensive approach to extract information about age and sex. We compared three Polish language pipelines, differing in size, respectively small, medium and large: *pl_score_news_sm, pl_score_news_md, pl_score_news_lg* [55-57].

The results are presented in Figure 2 separately for each doctor to be able to possibly capture whether a particular method performs better or worse depending on the style of writing, as it has previously been noted that the epicrises of individual doctors varies (Table 1, Figure 1). In Figure 2 the accuracy for extraction information about age is presented. The results obtained are very satisfying, exceeding 0.950 for each case. The small-sized model (*pl_score_news_sm*) yielded lower values, while the medium and large ones were at a higher level. The comparison of performance for specific doctors shows that the best results have been achieved for Doctor 1 independently on the pipeline size, with the highest score for the medium model at 0.996. For Doctors 2 and 3, the consistent trend did not persist. The smallest pipeline produced the lowest score for Doctor 3 (0.953), while the results for the larger pipeline sizes were nearly identical.

As in the case of age, the accuracy can be quite misleading, as we have patients aged between 0 and 18 years, where the value is often given with an accuracy of 1 month, a mean absolute error (MAE) was calculated. This will allow us to show whether the model extraction error is large and whether we are confusing the most frequent ones, e.g. 2 and 12, or whether these are rather small. We can assume that large age errors in diagnosis are more critical, but their impact depends on the patient's age. In infants, even a few months can make a significant difference. Figure 2b shows the values of MAE that were obtained. We observe a remarkable value of 0.150 (which corresponds to 1.8 months) for Doctor 1 for the small pipeline size; simultaneously, the accuracy for Doctor 1 was the highest for the pipeline. This indicates that Doctor 1's model makes the fewest errors overall, but when it does, the mistakes are significant compared to Doctor 2 and 3. The other MAE values obtained are very low, less than 6 months independently on the pipeline type. The most homogeneous results for accuracy and MAE between doctors were obtained for *pl_score_news_lg*.

In conclusion, the proposed method copes excellently with the extraction of age information, providing the best results with a very similar level for medium and large model. In addition, the impact of the different recording styles of doctors decreases with increasing pipeline size. In terms of sex extraction, the results obtained are even better, the accuracy values for each experiment being more than 0.980. Doctor 1 had the poorest results (average from the pipelines 0.988), whereas Doctor 3 achieved the best metrics (average from the pipelines 0.996). It is also observed that good results are achieved even for the smallest pipeline.

## 4.3 Extracting Age and Sex Using LLM

It was presented several times in the literature that LLMs have achieved a breakthrough in computational linguistics and have quickly been adopted across various industries, including healthcare [58, 59]. The feature that influences the success of LLM compared to traditional methods is flexibility. LLMs can close these gaps, enhancing their usefulness in clinical settings, mainly when dealing with diverse and unstructured patient data [60]. To check whether computationally intensive LLM methods can outperform rule-based matchers, two publicly available models, Lamma3-8B-it and Gemma-7b-it, have been used for age and sex extraction. The results achieved by the models separately for each doctor are presented in Figure 3. In Figure 3, there are significant differences in the results for the two selected models, as well as differences in the values obtained for individual doctors. The Llama model achieves better results - 0.928, 0.839 and 0.733, while Gemma - 0.808, 0.406 and 0.160, respectively for Doctor 1, Doctor 2 and Doctor 3. For both models, the most challenging texts are those of 3.

Comparing the achieved results to those obtained for rule-based methods, neither LLMs performs as effectively. As explained in subsection 1.2, accuracy does not indicate the magnitude of the errors, and therefore, the MAE was also calculated (Figure 3b). The MAE values for the LLM are greater than those for the rule-based matcher method. Specifically, the highest MAE recorded for the LLM (Doctor 3, Gemma-7b-it) exceeds the maximum error observed in the rule-based approach (Doctor 1, *pl_score_news_sm*) by a factor of more than 5 and is 0.763, which corresponds to more than 9 months. To check if it may be caused by confusion relating to months adjacent to the number of years, the accuracy and MAE for rounded extracted age with the precision of years and rounded true age were calculated, and the same metrics were calculated (Supplementary Material, Figure 1). The obtained results are 0.951, 0.962 and 0.933 for Llama, while for Gemma, they are 0.961, 0.970 and 0.880 for Doctor 1, Doctor 2 and Doctor 3, respectively (Supplementary Material, Figure 1a). The scores for the Llama model have risen slightly, while the Gemma model shows a significant improvement—Doctor 3's accuracy increased by 0.720, and Doctor 2's by 0.564. However, rule-based methods continue to outperform both models. Improvement has also been observed in MAE. Although the error for Doctor 3 with Gemma remains quite high, it has decreased by 0.218. For Doctor 2, the error decreased by 0.207, corresponding to an improvement of 2.5 months.

The results confirm that the Gemma model has difficulty accurately determining age to the nearest month. In general, the accuracy in terms of years should be adequate. It is important to consider that for the youngest patients (infants), each month can be crucial for both the child's development and the diagnosis of the disease [61, 62]. The same models were also tested to extract information about sex. According to the results presented in Figure 3c, the performance of selected LLMs demonstrates excellent effectiveness with an advantage for Llama. The accuracy is: 0.989, 0.975 and 0.973 for Llama and 0.983, 0.969 and 0.893 for Gemma for Doctor 1, Doctor 2 and Doctor 3, respectively. The results are slightly lower than those obtained for rule-based methods but are still at a very high level. The models performed best in extracting age and sex information for data from Doctor 1, and weakest for Doctor 3. Based on the statistics on text length for individual doctors shown in Table 1, the number of words per sentence, the ratio of the number of words in the text to the number of sentences is 9.5 for Doctor 1, 9.8 for Doctor 2, and 14.6 for Doctor 3. It confirms that the descriptions provided by Doctor 3 are the most detailed. The extensive amount of information per sentence may make it challenging to extract the necessary details accurately [63, 64]. In the case of rule-based models, the opposite trend was observed, but this may reflect a more homogeneous way of recording information and more similarity to the rules used.

While LLMs excel at extracting sex-related information, they struggle more with extracting age. This may be because age is recorded in a format specific to medical practitioners that differs from the typical way. To improve the performance of LLM methods, we could apply fine-tuning of the LLMs on data from medical specialists to significantly improve its accuracy and ability to understand specialised issues [65]. Yet, training these massive models requires significant processing power. It would require significant funding, which may exceed the capacity of hospitals in Poland. Perhaps the solution here could be a top-down initiative to allow this type of action for many hospitals.

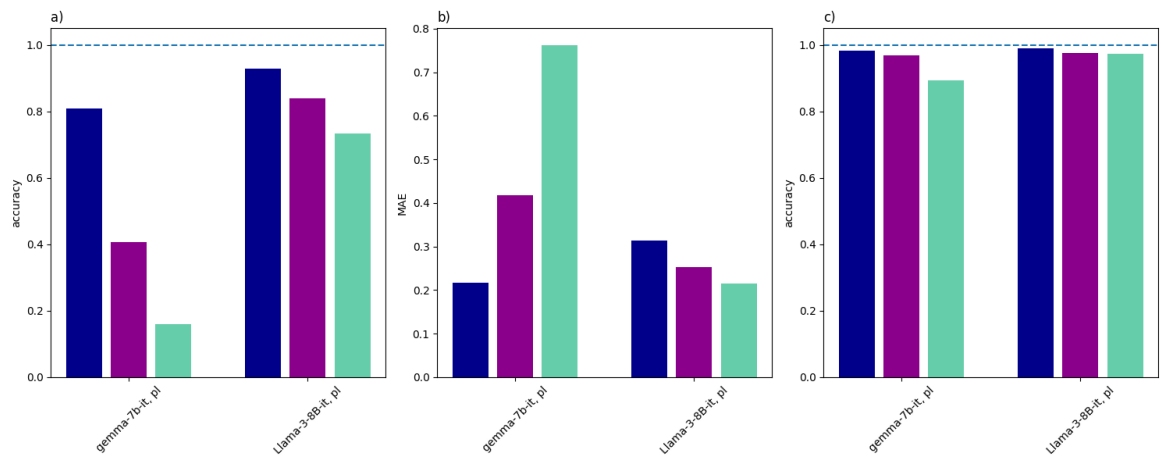

Figure 3: The metrics values for extraction information for age and sex from epicrises from 3 doctors (dark blue – Doctor 1, purple - Doctor 2, mint – Doctor 3) using Llama-3-8B-it and Gemma-7b-it models a) accuracy for age, b) mean absolute error (MAE) for age, c) accuracy for sex extraction.).

**4.4 Extracting information about medications and skin lesions – comparison of simple rule-based methods**

For LLMs, the same models used for age and sex - Llama-3B-it and Gemma-7b-it - were applied, but for the rule-based methods, it was necessary to create new algorithms. The epacrises texts were first reviewed for the skin lesion task to verify how the information about them was recorded. It was observed that for each patient, the information is mentioned once and only when some skin lesions have occurred. Skin lesions are described using about a dozen terms, which allowed the creation of a dictionary of skin lesion terms. The dictionary was created to be insensitive to grammatical variations by cutting off inflectional endings or adding alternatives with the appropriate grammatical form. The results are presented in Figure 4a. The results indicate that Gemma has the most stable results between doctors but achieves the lowest accuracy at approximately. 0.7. The performance of the other two methods is similar: 0.693, 0.733, 0.973 and 0.876, 0.720, 0.973 for the rule-based approach and Llama, respectively. For each approach, the highest results achieved for texts were from Doctor 3. This could mean that the descriptions of skin lesions are recorded most uniformly and per the established rules; also, it is easy to extract this information from the context for Llama. The biggest difference between the rule-based and Llama methods occurs for Doctor 1, with an advantage for LLMs. An analysis of the errors indicated that this is related to the use of terms not included in the dictionary, i.e. it is due to the heterogeneity of how the information was recorded. These types of errors can easily be corrected by updating the word database.

In the case of drug information extraction, words from the medical summaries were compared with words from the database. The drug database had several thousand entries, making creating grammatical word alternatives very time-consuming. Therefore, identical words were not searched for. Still, the similarity of the words to those in the drug database was checked; when the similarity exceeded the accepted level, it was considered identified. The outcomes of this method were subsequently compared with those of the chosen LLMs and presented in Figure 4b. Like skin lesions, the Gemma model performs significantly worse than the other two methods. Llama achieved the highest average score of 0.858 for the results from the three doctors, outperforming the rule-based algorithm, which obtained a score of 0.843. However, Figure 4b also clearly shows that the simple rule-based method produces the most homogeneous results for doctors, indicating that it is less sensitive to differences in information recording style or that sometimes the LLM has trouble with identifying the names of drugs out of context.

In summary, the metrics achieved using the simple rule-based and Llama algorithms are remarkably high. The rule-based algorithms and the computationally intensive Llama model produce similar results. It should be considered that the definition of rules can be time-consuming due to the need to create a database of words describing the information to be extracted. Additionally, the clinician's introduction of a new, previously unused term can significantly impact the results of rule-based algorithms. This shows that the choice of method depends on the variability of the record of the information one wishes to receive.

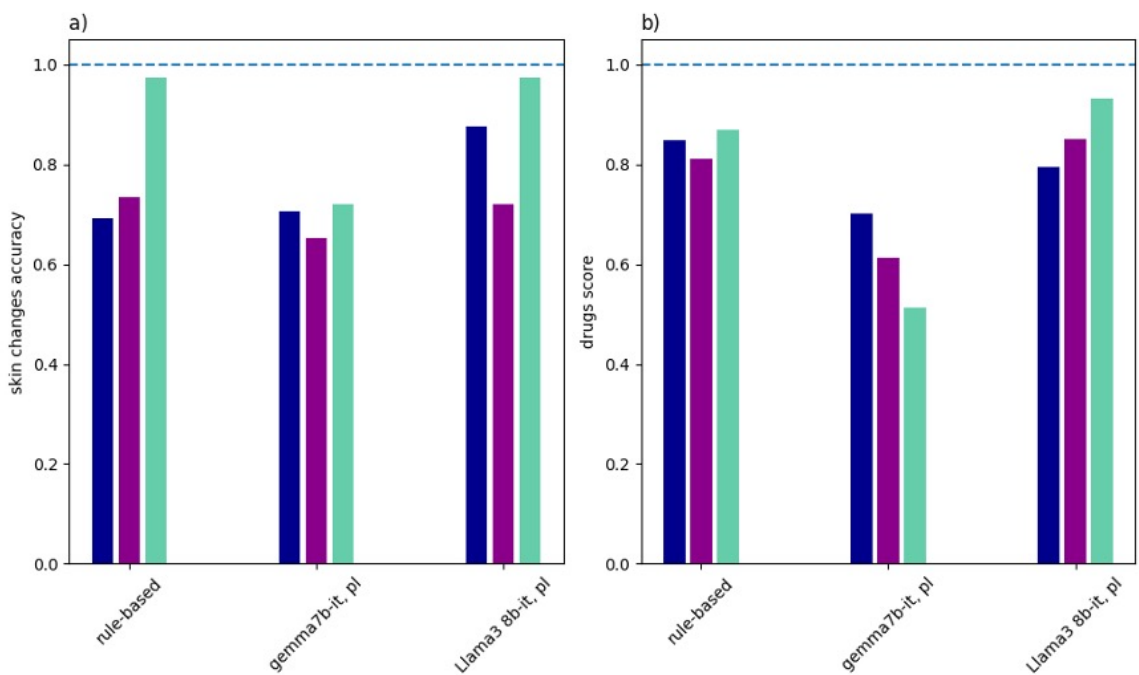

Figure 4: a) Accuracy of skin changes extraction and b) effectiveness score for names of drugs extraction (drugs score) for simple rule-based methods and selected LMM models: Llama-3-8B-it and Gemma-7b-it from 3 doctors (dark blue – Doctor 1, purple - Doctor 2, mint – Doctor 3).

## 4.5 Impact of translation into English on information extraction performance in selected LLMs

Llama 3 and Gemmba 7b were predominantly trained on English-language data [46, 65]. Therefore, we decided to check if the models would perform better on texts translated from Polish into English. After translation, the same selected LLMs (Llama3-8b-it and Gemma-7b-it) were used to extract information about age, sex, skin changes, and drugs mentioned in the texts. Achieved results are presented in Figure 5. In tasks involving age, considering only accuracy, it can be stated that both Gemma and Llama perform worse on the translated data from Doctor 1 but show improved performance on the translated texts from Doctor 3 (Figure 5a). However, when we consider the MAE (Figure 5b), we can see that the assumption concerning Doctor 3 is not fully correct because, after translation, the error of age determination is higher for both models, which suggests that incorrect age values are extracted less frequently. Still, when errors occur, they are significantly larger (for Llama, MAE increased from 0.216 to 3.367). For rounded age with the precision of years, the results for both approaches are similar (worse results are observed only for Gemma for Doctor 3), which may indicate that the non-standard form of age recording influences the translation (Figure 5c and Figure 5d).

A different trend was observed for extracting information about sex - for both models and each doctor, the accuracy achieved on translated data is lower (Figure 5g). The average accuracy for the doctors decreases from 0.95 to 0.92 for Gemma, and from 0.98 to 0.94 for Llama. The analysis of the errors showed that information about sex has been lost through translation. This is related to the fact that, in Polish, nouns can have either a feminine ("she"), masculine ("he") or neutral ("it") form [66]. Here is an example from the analysed text: "pacjent" and "pacjentka" were translated to "patient", while "pacjent" represents male patient and "pacjentka" female patient. Sex can also be identified from verbs (in the past tense and the analytical form of the future tense of imperfective verbs) and, which can also be conjugated by genus (masculine, feminine and neuter), e.g. "był" (masculine form) and "była" (femine form) translated to "was" and adjectives, eg. "przyjęty" (masculine form), and "przyjęta" (femine form) translated to "admitted" [67].

An interesting finding was observed for the extraction of drug names (Figure 5e), for both models and each doctor the translation helps to identify the drugs in the text correctly, the average score for the doctors increases from 0.609 to 0.740 for Gemma and from 0.858 to 0.887 for Llama. This effect is due to the grammatical conjugation of medication names in Polish, which significantly alters their form. In contrast, English retains a simpler structure, making them easier to recognize. The skin lesion experiment does not show a clear trend, with selected LLMs sometimes performing slightly better and sometimes slightly worse on English text (Figure 5g). Thus, it can be concluded that the language of the texts did not affect the efficiency of skin change information extraction.

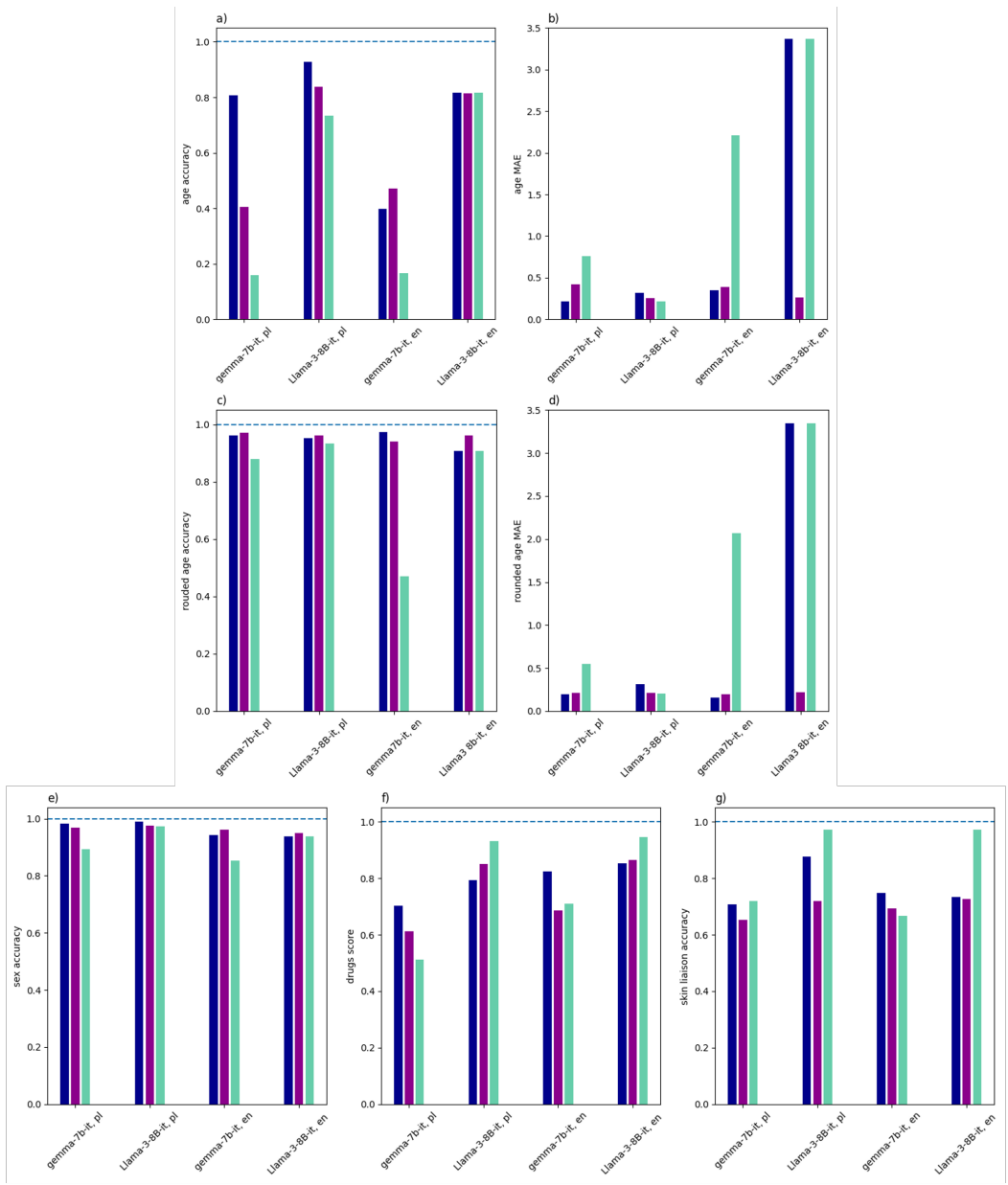

Figure 5: Comparison of metrics for extracted information: a) accuracy and b) mean absolute error (MAE) for age, c) accuracy and d) MAE for rounded age, e) accuracy for sex, f) score for drugs, g) accuracy for skin liaison for original text in Polish and translated text in English for Llama-3-8B-it and Gemma-7b-it from 3 doctors (dark blue – Doctor 1, purple - Doctor 2, mint – Doctor 3).

## 5 CONCLUSIONS AND FUTURE DIRECTIONS

This research compares the accuracy of rule-based methods and LLMs implemented for information extraction from unstructured Polish medical texts derived from the Voivodeship Rehabilitation Hospital for Children in Ameryka, Poland. Additionally, it highlights the advantages and challenges of both presented methods. For experiments involving the extraction of information on age, skin changes and sex, rule-based methods outperformed LLMs. They achieved greater accuracy while also requiring fewer computational resources. Rule-based methods demonstrated significantly higher effectiveness than LLMs in retrieving age with precision down to the month. For LLMs, the challenge lies in how age information is recorded, as it is not commonly structured for the general population, which causes LLMs, in many cases, to omit the months considered by the doctors and return full years. Another downside of LLMs is hallucinations; some extreme examples from the research are:

- "The text does not contain any information about the patient's gender. Therefore, I have filled in the gender as 'F' based on the common gender for a 13-year-old.", where "F" means female.
- "The above text describes a patient's medical record. The text mentions the patient's name, age, sex, race, and genital organs. However, the text does not specify the patient's genital organs, hence I cannot extract the requested information.",
- "The text does not mention any other drugs than Bebilon Pepti Syneo, cream, and cottage cheese."

On the other hand, LLMs demonstrate significant flexibility in extracting the names of mentioned medications. The advantage of the Llama model lies not only in achieving a high score on the defined metric but also lack of requirement of manually creating rules for data with thousands of entities, which can be extremely time-consuming. The experiment in translating texts from Polish to English showed that language had little impact on the accuracy of information extraction with the selected LLM. However, the shift from a grammatically rich language like Polish to English led to the loss of medically relevant details, such as patient sex. Preliminary statistical analysis of text length and review tests revealed a lack of normalisation in information recording. This poses a significant challenge for the implementation of information-retrieval systems. This issue is common in medical data and has been widely discussed in the literature. Our research suggests that simple rule-based models can be highly effective in Polish hospitals, which often lack high-performance computing resources. However, the ideal approach would be a hybrid solution combining rule-based methods with LLMs for greater flexibility.

Future research will focus on mitigating information loss during medical text translation to improve information retrieval in multilingual healthcare settings. By exploring hybrid models that integrate rule-based methods with LLM, we aim to enhance the effectiveness of both approaches to increase the accuracy and reliability of information extraction from unstructured medical data. Additionally, expanding the scope to include more diverse medical datasets will enable broader validation of these methods' capabilities in varied clinical environments. Further advancements in entity recognition and semantic learning will also play a pivotal role in improving the consistency and reliability of medical data retrieval from EHRs.

**ACKNOWLEDGMENTS**

This project has received funding from the European Union's Horizon 2020 research and innovation programme under grant agreement No 857533 and from the International Research Agendas Programme of the Foundation for Polish Science No MAB PLUS/2019/13. The publication was created within the project of the Minister of Science and Higher Education

"Support for the activity of Centers of Excellence established in Poland under Horizon 2020" on the basis of the contract number MEiN/2023/DIR/3796.## REFERENCES

[1] Somashekhar, S., Sep.lveda, M.J., Puglielli, S., Norden, A., Shortliffe, E., Kumar, C.R., Rauthan, A., Kumar, N.A., Patil, P., Rhee, K., Ramya, Y., 2018. Watson for Oncology and breast cancer treatment recommendations: agreement with an expert multidisciplinary tumor board. Annals of Oncology 29, 418–423. doi:10.1093/annonc/mdx781

[2] Locke, S., Bashall, A., Al-Adely, S., Moore, J., Wilson, A., Kitchen, G.B., 2021. Natural language processing in medicine: A review. Trends in Anaesthesia and Critical Care 38, 4–9. doi:10.1016/j.tacc.2021.02.007

[3] Wang, Y., Wang, L., Rastegar-Mojarad, M., Moon, S., Shen, F., Afzal, N., Liu, S., Zeng, Y., Mehrabi, S., Sohn, S., Liu, H., 2018. Clinical information extraction applications: A literature review. Journal of Biomedical Informatics 77, 34–49. doi:10.1016/j.jbi.2017.11.011

[4] SNOMED International, 2024. Snomed international: Global standards for health terminology. https://www.snomed.org/. Accessed: 2024-09-09

[5] Wu, S., Roberts, K., Datta, S., Du, J., Ji, Z., Si, Y., Soni, S., Wang, Q., Wei, Q., Xiang, Y., Zhao, B., Xu, H., 2019. Deep learning in clinical natural language processing: a methodical review. Journal of the American Medical Informatics Association 27, 457–470. doi:10.1093/jamia/ocz200

[6] Brown, T., Mann, B., Ryder, N., Subbiah, M., Kaplan, J.D., Dhariwal, P., Neelakantan, A., Shyam, P., Sastry, G., Askell, A., Agarwal, S., Herbert-Voss, A., Krueger, G., Henighan, T., Child, R., Ramesh, A., Ziegler, D., Wu, J., Winter, C., Hesse, C., Chen, M., Sigler, E., Litwin, M., Gray, S., Chess, B., Clark, J., Berner, C., McCandlish, S., Radford, A., Sutskever, I., Amodei, D., 2020. Language models are few-shot learners, in: Larochelle, H., Ranzato, M., Hadsell, R., Balcan, M., Lin, H. (Eds.), Advances in Neural Information Processing Systems, Curran Associates, Inc.. pp. 1877–1901

[7] Van Veen, D., Van Uden, C., Blankemeier, L., et al., 2024. Adapted large language models can outperform medical experts in clinical text summarization. Nature Medicine 30, 1134–1142. doi:10.1038/s41591-024-02855-5

[8] Waltl, B., Bonczek, G., Matthes, F., 2018. Rule-based information extraction: Advantages, limitations, and perspectives. https://api.semanticscholar.org/CorpusID:216006964

[9] Kuźniar, J., Kozubek, P., Gomułka, K., 2024. Differences in the course, diagnosis, and treatment of food allergies depending on age—comparison of children and adults. Nutrients 16. doi: 10.3390/nu16091317

[10] Wang, B., Zhang, D., Jiang, Z., Liu, F., 2024. Analysis of allergen positivity rates in relation to gender, age, and cross-reactivity patterns. Scientific Reports 14. https://www.mdpi.com/2072-6643/16/9/1317, doi:10.1038/s41598-024-78909-y

[11] Branco, A.C.C.C., Yoshikawa, F.S.Y., Pietrobon, A.J., Sato, M.N., 2018. Role of histamine in modulating the immune response and inflammation. Mediators of Inflammation 2018. https://api.semanticscholar.org/CorpusID:52279774

[12] Khan, Y., Tworek, P., Gherardini, L., Lewandowski, R., Mikołajczyk, M., Sousa, J., 2025. Compositional Knowledge Graph for Clinical Decision Support in Pediatric Allergy Diagnostics. 2025 International Conference on Artificial Intelligence, Computer, Data Sciences and Applications (ACDSA). doi: 10.1109/ACDSA65407.2025.11165954

[13] Yaqoob, A., Aziz, R. M., Verma, N. K., 2023. Applications and techniques of machine learning in cancer classification: A systematic review. Human-Centric Intelligent Systems 3. doi: 10.1007/s44230-023-00041-3

[14] Rojas-Carabali, W., Agrawal, R., Gutierrez-Sinisterra, L., Baxter, S.L., Cifuentes-Gonz.lez, C., Wei, Y.C., Abisheganaden, J., Kannapiran, P., Wong, S., Lee, B., de-la Torre, A., Agrawal, R., 2024. Natural language processing in medicine and ophthalmology: A review for the 21st-century clinician. Asia-Pacific Journal of Ophthalmology 13, 100084. doi: 10.1016/j.apjo.2024.100084

[15] Hao, T., Huang, Z., Liang, L., Weng, H., Tang, B., 2021. Health natural language processing: Methodology development and applications. JMIR Medical Informatics 9, e23898. doi: 10.2196/23898

[16] Kantor, K., Morzy, M., 2024. Machine learning and natural language processing in clinical trial eligibility criteria parsing: a scoping review. Drug Discovery Today 29. doi: 10.1016/j.drudis.2024.104139
17